\documentclass{article}

\usepackage{arxiv}

\usepackage[utf8]{inputenc} 
\usepackage[T1]{fontenc}    
\usepackage{hyperref}       
\usepackage{url}            
\usepackage{booktabs}       
\usepackage{amsfonts}       
\usepackage{nicefrac}       
\usepackage{microtype}      
\usepackage{graphicx}
\usepackage{natbib}
\usepackage{multirow}

\title{Evaluating Large Language Models for Causal Modeling}

\author{ \href{https://orcid.org/0000-0002-9527-5781}{\includegraphics[scale=0.02]{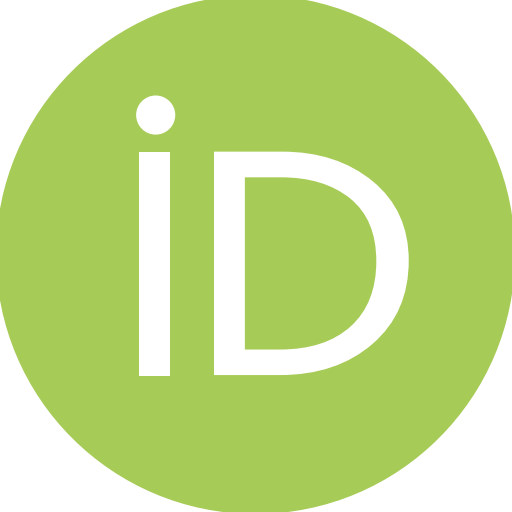}\hspace{1mm}Houssam Razouk}\thanks{Corresponding Author: \texttt{Houssam.razouk@student.tugraz.at} } \\
	Graz University of Technology, Graz, Austria\\
	\And
	\href{https://orcid.org/0009-0005-5841-8594}{\includegraphics[scale=0.02]{orcid.png}\hspace{1mm}Leonie Benischke} \\
	Université Bourgogne Franche-Comté, Besançon, Franc\\
        \And
	\href{https://orcid.org/0009-0008-9569-1313}{\includegraphics[scale=0.02]{orcid.png}\hspace{1mm}Georg Niess} \\
	Graz University of Technology, Graz, Austria\\
        \And
        \href{https://orcid.org/0000-0003-0202-6100}{\includegraphics[scale=0.02]{orcid.png}\hspace{1mm}Roman Kern} \\
	Graz University of Technology, Graz, Austria\\
        Know-Center GmbH, Graz, Austria \\
}

\begin{document}
\maketitle

\begin{abstract}
In this paper, we consider the process of transforming causal domain knowledge into a representation that aligns more closely with guidelines from causal data science.
To this end, we introduce two novel tasks related to distilling causal domain knowledge into causal variables and detecting interaction entities using LLMs.
We have determined that contemporary LLMs are helpful tools for conducting causal modeling tasks in collaboration with human experts, as they can provide a wider perspective.
Specifically, LLMs, such as GPT-4-turbo and Llama3-70b, perform better in distilling causal domain knowledge into causal variables compared to sparse expert models, such as Mixtral-8×22b.
On the contrary, sparse expert models such as Mixtral-8×22b stand out as the most effective in identifying interaction entities.
Finally, we highlight the dependency between the domain where the entities are generated and the performance of the chosen LLM for causal modeling\footnote{Code 
 and data are available under https://anonymous.4open.science/r/expirment-using-LLMS-for-causal-modeling-254E/readme.md}
\end{abstract}


\section{Introduction}

\begin{figure}
\centering
  \includegraphics[width=0.5\columnwidth]{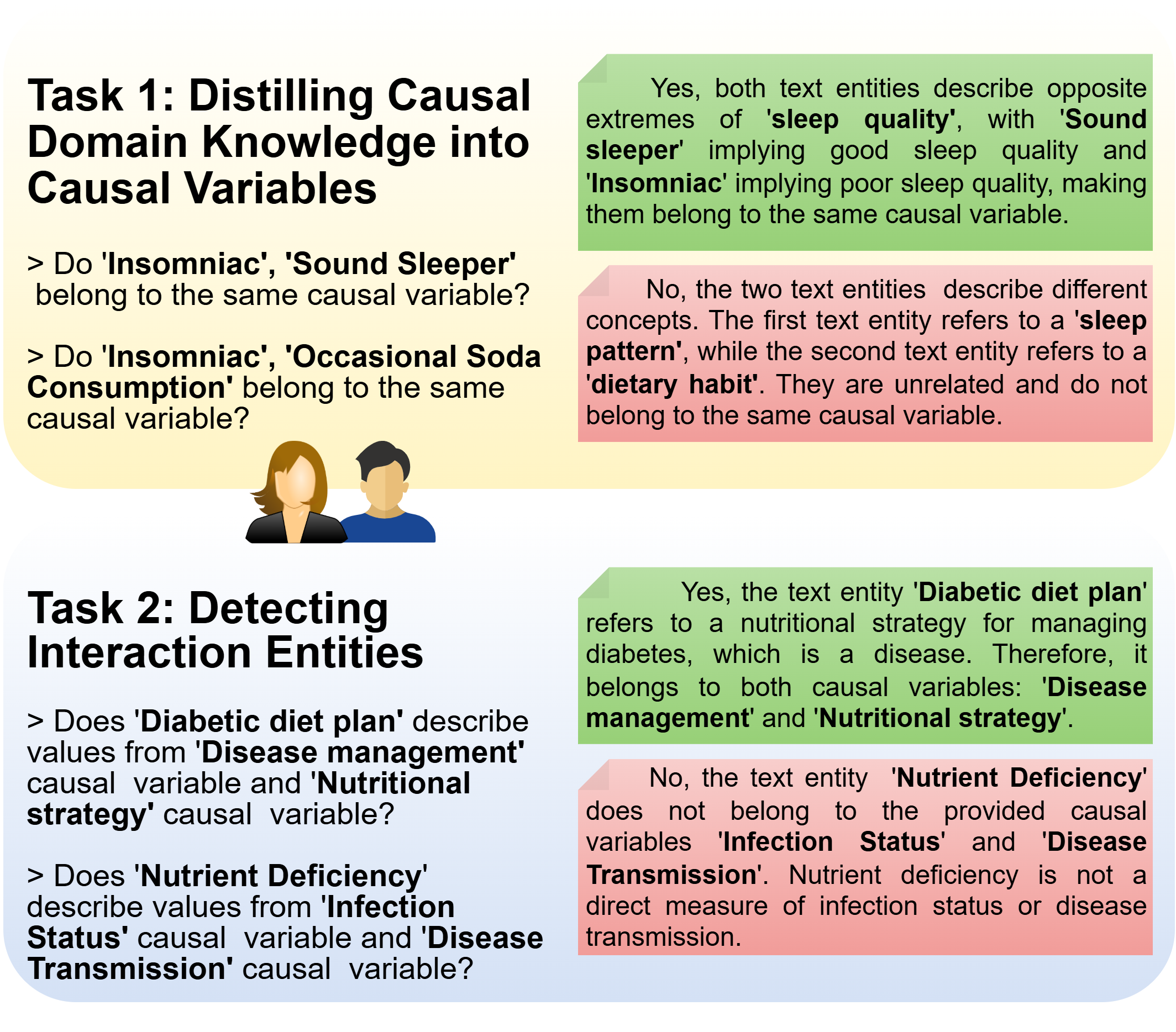}
  \caption{\textbf{Causal modeling of domain knowledge by human experts.}
  This is a tedious and error prone process.
  We study the usefulness of LLMs to support this process, whereby we first divide this into two distinct tasks.
  }
  \label{fig:fig1}
\end{figure}

Causal modeling plays a crucial role in driving reliable and robust data analysis which is resource intensive and error prone.
Causal modeling is responsible for distilling and representing causal domain knowledge in the form of causal variables and relations.
Specifically, causal modeling includes distilling causal domain knowledge into causal variables and detecting interaction entities, as depicted in Figure~\ref{fig:fig1}.

One great example of the effective use of causal domain knowledge is illustrated in addressing the Simpson's paradox~\citep{pearl2011simpson}.
In this example, \cite{pearl2011simpson} illustrates that the paradox resolution relies on the fact that causality is governed by its own logic and cannot be resolved with observational data alone.
In many cases, causal domain knowledge, which is used in causal modeling, is collected through brainstorming approaches or is extracted from textual documents using causal information extraction techniques.
For example, causal domain knowledge is collected through brainstorming approaches  using cognitive mapping approaches~\citep{pinto2023analyzing,faria2018combining}.
Also,  causal domain knowledge is extracted from textual documents using causal information extraction techniques for constructing causal graphs~\citep{gopalakrishnan2023text, hashimoto2019weakly}.
However, this causal domain knowledge is often expressed as text entities that describe realized values instead of the causal variables it represents.

For instance, considering the two entities "Insomniac" and "Sound Sleeper", they describe different values of the causal variable "Sleep quality". 
Representing these entities as independent causal variables in a causal model has several drawbacks, including violating the modality of the causal model in the variables~\citep{Kuorikoski,suzuki2020causal}.
Also, considering the interaction entity "Diabetic diet plan", this entity describes the value "Diabetes management" from the causal variable "Disease management" and the value "diet plan" from the causal variable "Nutritional strategy".
Representing this value as part of the causal variables "Diabetes management" or "Disease management" could comprise the transitivity principle of the causal relation~\citep{mcdonnell2018transitivity}.

Ideally, it is recommended to group these entities under the causal variable it represents~\citep{suzuki2020causal}.
In this example, it is recommended to group "Insomniac" and "Sound Sleeper" under "Sleep quality".
Also, it is recommended  to represent interaction entities as artificial variables explicitly indicating that these entities are the results of interaction between multiple causal variables similar to work presented in~\cite{VanderWeele_2009,mcdonnell2018transitivity}. 
In this example, it is recommended to represent "Diabetic diet plan" as artificial variables explicitly indicating the interaction between the causal variables "Disease management" and "Nutritional strategy".
However, it is widely acknowledged that extracting causal models from this causal domain knowledge requires significant resources\citep{pinto2023analyzing} and is prone to errors\citep{mcdonnell2018transitivity,suzuki2020causal,Kuorikoski}.

At the same time, research from the field of causality is increasingly focused on testing the capabilities of large language models (LLMs)~\citep{jin2023can,zevcevic2023causal}.
For example, \citealp{jin2023can} generated causal diagrams and examined the ability of LLMs to identify various types of relationships between variables. 
Furthermore, \citealp{zevcevic2023causal} argue that LLMs do not understand the mechanization of causal relations but rather echo knowledge based on their training domain.
However, few researchers have systematically tested LLMs ability to distill causes and effects from natural language into causal variables.

This paper presents our experimental investigation into the capability of LLMs to distill causal relationships described between text entities into specific causal variables. 
In particular, we examine two scenarios. 
Firstly, we assess the capacity of the LLMs to determine if two given entities, each portraying a cause or an effect, represent different values of the same causal variable without explicitly stating the causal variable itself. 
Secondly, we analyze a scenario in which an entity is provided along with a set of causal variables (two or more), and we assess whether the entity represents an interaction between the given causal variables.

Our experiment follows a within-subjects design, encompassing various LLMs and diverse domains.
Each LLM is tasked with producing text entities related to the values of causal variables and their interaction within each domain. The generated data is sampled to create an evaluation dataset comprising positive and negative samples from each model.
Subsequently, each model is tasked with generating predictions over the generated dataset.
Cohen's $kappa$ coefficient~\citep{cohen1960coefficient} is employed to estimate the agreement between the predictions of the LLM and the generated data of that LLM and other LLMs under study.

The present study contributes to our understanding of LLMs causal modeling abilities. 
Specifically, this study contributes by introducing two novel tasks focused on transforming causal domain knowledge into a representation that aligns more closely with guidelines from causal data science.
It highlights the potential of LLMs to expedite current approaches used for causal modeling, as these approaches are resource intensive and error prone, thereby streamlining the process of selecting the most suitable LLM for a given domain.
This study illuminates LLMs current limitations and paves the way for including these tasks in future LLM benchmarks.

\section{Related Work}

A comprehensive review of the relevant literature related to the modeling of causal domain knowledge is provided in Section~\ref{Causal modeling related work}.
Furthermore, recent work on extracting causal information from text is summarized in Section~\ref{Causal information extraction from text}. 
Finally, work on testing the causal capabilities of LLMs is explored in Section~\ref{Testing causal modeling abilities of LLMS}.

\subsection{Causal Domain Knowledge Modeling Related Work}
\label{Causal modeling related work}

Causal modeling plays a crucial role in driving reliable and robust data analysis.
As such, causal data science is gaining more interest in parallel to the rapid development of data-driven approaches, such as machine learning algorithms and the area of big data~\citep{Guo_2020}.
In machine learning, the prediction quality is highly dependent on the validity of the assumptions made during the data analytic process.
Greenland summarized the most common causal frameworks in his work~\citep{greenland2002overview}.

However, causal modeling in real-world scenarios, for example, using causal graphical models, can be challenging due to potential pitfalls that can easily be overlooked~\citep{suzuki2020causal}. 
To narrow this gap, \citealp{suzuki2020causal} provide a set of guidelines for scholars who are using causal diagrams, particularly directed acyclic graphs (DAGs). 
These guidelines include distilling the causal domain knowledge into causal variables in the causal model, not its realized value. 
This approach ensures that the causal models represent the underlying causal mechanisms rather than just the variables observable values.

At the same time, to ensure the validity of causal models, it is crucial to consider the principles of transitivity and proportionality in causation, as outlined in Neil McDonnell's work~\citep{mcdonnell2018transitivity}. 
Transitivity is a logical principle that states that if A causes B and B causes C, then A must also cause C. 
This principle is often used as a criterion to check whether the relationships between variables are consistent. 
When transitivity is violated, it may indicate that the causal model needs to be revised.
Specifically,~\cite{mcdonnell2018transitivity} illustrated that the transitivity principle of the causal relation could be violated in the case of interaction entities.

In addition, the concept of modularity, as it applies to causal diagrams, is discussed by Kuorikoski~\citep{Kuorikoski}.
Modularity is important since it allows different parts of a causal diagram to be separated and analyzed independently.
Kuorikoski distinguishes between three types of modularity in causal diagrams.
Firstly, there is causal model modularity in variables. 
This refers to surgically intervening on a variable (i.e., assigning a value to the variable) in the causal model without affecting other variables.
Secondly, there is variable mechanism\footnote{The causal effect mechanism indicates how the treatment affects the outcome} modularity.
This indicates the independence between the variable (or at least a range of the variable) and the causal effect mechanism. 
This ensures the smoothness of the outcome, which means that the causal mechanism is independent of the variable values.
Finally, there is causal model modularity in parameters\footnote{Parameters are the variables which are assumed to be constant when developing the causal model for a specific context}. 
This refers to the ability to intervene on a parameter in the causal model and affect only one causal effect relation.
Overall, modularity allows for a more detailed and nuanced analysis of causal relationships in a system by breaking down the system into its constituent parts and analyzing them independently.

More recently, \citealp{zhang2022causal} differentiate between explicit and generic variables. An explicit variable can be viewed as a specialized version of a generic variable as it pertains to a specific unit, sample, or individual.
While a generic variable, such as "X", applies to a range of instances, an explicit variable, such as "unit i (Xi)" is tailored to a specific unit and its corresponding attribute or event.
With this differentiation, \citealp{zhang2022causal} were able to propose definitions of the interaction model and the isolated interaction model.
These definitions are necessary for more reliable results in settings where the IID (Independent Identical Distribution) assumption does not hold.

In short, although the principles of causal modeling have existed for some time ~\citep{greenland2002overview,suzuki2020causal,mcdonnell2018transitivity,Kuorikoski,zhang2022causal}, they have not been widely adopted. 
This assertion is supported by our observations on the application of causal analysis to the results of cognitive maps in works related to urban studies~\citep{pinto2023analyzing,pinto2021analyzing,ferreira2016you,braga2021dematel,ribeiro2017fuzzy}, finance~\citep{rodrigues2022artificial,barroso2019analyzing}, psychology~\citep{ferreira2015operationalizing}.
Specifically, cognitive maps encompass textual entities that describe the causes of an effect in a causal relation.
In many cases, these textual entities describe the values of the causal variables and the variables themselves. 
Furthermore, these textual entities could describe the interaction between causal variables, as highlighted in the work by~\cite{razouk2022improving}.

\subsection{Causal Information Extraction from Text}
\label{Causal information extraction from text}

Extracting causal information from text is a topic of great interest since not only numerical data but also texts can provide valuable insight into causal relations~\citep{yuan2023tc}. 
Although causal discovery aims to uncover the causal model or, at the very least, a Markov equivalent that mimics the data generation process for a given data set~\citep{glymour2019review}, causal information extraction from texts focuses on identifying causal entities and how they are connected to each other~\citep{yang2022survey}.

Some examples of causal information extraction can be reviewed in the work by ~\cite{saha2022spock}. 
Specifically, they label the text spans corresponding to cause and effect in a given text. 
They proceed to classify whether these identified cause and effect spans are linked through a causal relation. 
Similarly,~\cite{khetan2020causal} employ an event-aware language model to predict causal relations by considering event information, sentence context, and masked event context. 
Another significant difficulty in extracting causality is the recognition of overlapping and nested entities. 
In response,~\cite{lee2022mnlp} tackle overlapping entities by employing the Text-to-Text Transformer (T5).
In addition,~\cite{gaerber2022causal} has proposed a multistage sequence tagging (MST) approach to extract causal information from historical texts. 
The MST method extracts causal cues in the first stage and then uses this information to extract complete causal relations in subsequent stages.
More recent work presented by \citealp{liu2023event} proposes an implicit cause-effect interaction framework to improve the reasoning ability of the model, which tackles event causality extraction generatively using LLMs. 

To summarise, extracting causal information from text is also moving toward leveraging the capabilities of large language models (LLMs). 
However, related research has systematically overlooked the testing of the abilities of LLMs in constructing causal models, which is a natural progression following causal information extraction.

\subsection{Testing Causal Modeling Abilities of LLMs}
\label{Testing causal modeling abilities of LLMS}

Most of the research on testing the causal modeling abilities of LLMs to date is centered around causal relations. 
For instance,~\cite{kiciman2023causal}  investigate the causal reasoning capabilities of large language models (LLMs) and highlight their potential to improve causal analysis across domains such as medicine, science, law, and politics. 
Their work demonstrates that LLMs could excel in pairwise cause discovery, counterfactual reasoning, and determining necessary and sufficient causes, often exceeding traditional algorithms. Furthermore, the study emphasizes the unique capacity of LLMs for natural language based causal graph generation and context identification, proposing their integration with existing methods to improve the efficiency and accuracy of high-level causal analysis.

At the same time,~\cite{he2023lego} introduce a multi-agent collaborative framework for generating causality explanation. 
Furthermore,~\cite{tucausal} test ChatGPT in causal discovery settings. 
Also,~\cite{jin2023can} scrutinize the ability of LLMs to infer causality from correlation and discover significant limitations, noting that LLMs often struggle to identify actual causal relationships instead of reproducing knowledge merely based on their training data.
Furthermore, ~\cite{zevcevic2023causal} argue that LLMs cannot perform causal inference as they lack an understanding of causal mechanisms and instead recite the causal domain knowledge embedded in their training data. 
Lastly,~\cite{li2023causal} focus on evaluating the causal reasoning ability of LLMs across different dimensions, such as causality identification and matching, to improve their reliability and applicability in practical scenarios.

Although most research has traditionally focused on examining causal reasoning, this paper examines explicitly causal variable modeling by testing the LLMs ability to determine (1) whether two text entities describe the values of the same variable or (2) whether a text entity describes the values of different causal variables.
We believe that causal reasoning often follows causal modeling, which is implicit in many cases when addressing a causal question. 
Causal modeling is of significant importance to transfer domain knowledge into a representation that aligns more with the recommendations for causal modeling outlined in Section~\ref{Causal modeling related work}.

\section{Method}

For this research, we have opted to assess the causal modeling capabilities of several LLMs across various domains in a zero-shot setting.
This decision is primarily influenced by the fact 
that we are interested in testing different LLMs innate causal modeling capabilities rather than their receptiveness to fine-tuning.
More information on problem formulation can be found in Section~\ref{Problem formulation}, while details on experimental design are provided in Section~\ref{Experiment design}.

\subsection{Problem Formulation}
\label{Problem formulation}

\textbf{Preliminaries:} 
Given a causal model $C_M$=($\mathcal{V}$, $\mathcal{M}$) with a set of causal variables $\mathcal{V}$ and a set of causal mechanisms~$\mathcal{M}$.
Each causal variable $V_i$~$\in$~$\mathcal{V}$ can take values $v_{i,j}$~$\in$~\{$v_{i,1}$,$v_{i,2}$,\dots,$v_{i,n}$\}.
Each causal mechanism $M$~$\in$~$\mathcal{M}$ encapsulates a causal path between two causal variables $V_c$~$\in$~$\mathcal{V}$ and $V_e$~$\in$~$\mathcal{V}$.
$V_c$ is a causal variable that acts as a cause or a treatment with respect to a causal mechanism $M$.
$V_e$ is a causal variable that acts as an effect or an outcome with respect to a causal mechanism $M$.
At the same time, given causal relations described in natural language $C_R$=($E$,~$R$)
with $E$: a set of entities that act as a cause or as an effect and $R$: a set of relations with causal semantics. 
Each entity $e$~$\in$~$E$ represents at least one value $v_{i,j}$ of a causal variable $V_i$~$\in$~$\mathcal{V}$.
Each causal relation is formed of a cause and an effect ($c$,~$e$)~$\in$~$R$~$\subseteq$~$E$~×~$E$. 
Each causal relation should be encapsulated within at least one causal mechanism  ($V_c$,~$M$,~$V_e$)~$\subseteq$~$\mathcal{V}$~×~$\mathcal{M}$~×~$\mathcal{V}$.

\subsubsection*{Task~1: Distilling Causal Domain Knowledge into Causal Variables} 
Causal models are primarily concerned about causal relations among variables~\citep{suzuki2020causal}. 
As such, we refer to random variables included in a causal model as causal variables.
One of the causal modeling targets is to distill causal domain knowledge included in $C_R$ into a set of causal variables~$\mathcal{V}$.
A simple yet effective approach to distilling causal domain knowledge into causal variables is to identify whether an entity $e_i$~$\in$~$E$ and an entity $e_j$~$\in$~$E$ represent different values of the same causal variable.

For example, given the two text entities "Insomniac" and "Sound sleeper",
human experts are able to state, based on their domain knowledge, that both entities describe opposite extremes of "Sleep quality", with "Sound sleeper" implying good sleep quality and "Insomniac" implying poor sleep quality, making them belong to the same causal variable.
Also, given the two text entities  "Insomniac" and "Occasional Soda Consumption", human experts are able to argue that the two entities describe different concepts. 
The first text refers to a "sleep pattern", while the second text refers to a "dietary habit". 
Thus, these entities do not belong to the same causal variable.

\subsubsection*{Task~2: Detecting Interaction Entities}
Interaction entities are text entities that act as a cause or an effect and describe values belonging to different causal variables at the same time.
Including these entities in the causal model as independent variables or as values under the causal variables it represents violates the modularity of causal variables of the model and the transitivity principle as shown by~\cite{mcdonnell2018transitivity,Kuorikoski,suzuki2020causal}.
Instead, it is recommended to include these interaction entities as artificial variables following the approach parented in \citep{VanderWeele_2009}.

It is sufficient for an entity to be an interaction entity if it represents values of more than one causal variable simultaneously.
A simple approach to detect if an entity $e$~$\in$~$E$ is indeed an interaction entity is to identify whether the entity $e$ represents values of the different causal variables \{$V_i$,\dots,~$V_k$\}~$\subseteq$~$\mathcal{V}$.  

For example, given the text entity "Diabetic diet plan" and the two causal variables "Disease management" and "Nutritional strategy",
human experts are able to state, based on their domain knowledge, that the entity refers to a nutritional strategy for managing diabetes, which is a disease. Therefore, it belongs to both causal variables.
Also, given the text entity  "Nutrient Deficiency" and the two causal variables "Infection Status" and "Disease Transmission",
human experts are able to argue that the text entity  "Nutrient Deficiency" does not belong to the provided causal variables. 
Nutrient deficiency is not a direct measure of infection status or disease transmission.

\subsection{Experiment Design}
\label{Experiment design}

\begin{figure*}
  \includegraphics[width=0.99\linewidth]{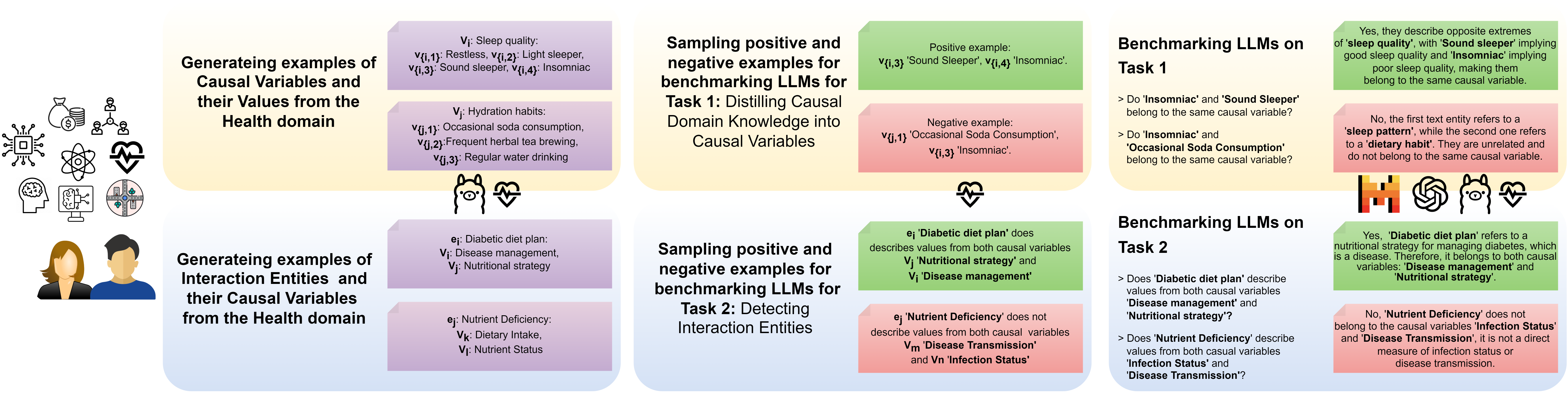} 
  \caption {\textbf{Experiments workflow on benchmarking LLMs causal modeling abilities.}
  Each LLM is instructed to generate data sets according to the task and the domain.
  The generated data sets are sampled to construct positive and negative examples.
  Each LLM is instructed to evaluate positive and negative examples sampled from its own and other LLMs generated data in a zero-shot setting.}
  \label{fig:fig2}
\end{figure*}

As depicted in Figure~\ref{fig:fig2}, our experiments use a within-subjects design involving all participating LLMs in every stage.
Each LLM generates two datasets and evaluates its own generated data and that of other LLMs.
The same prompts are utilized for all LLMs with a Temperature parameter of 0.
A text embedding model is devised to extract embeddings of the generated data. 
A classification method based on the cosine similarity threshold is employed to provide insights into the generated data and the LLMs predictions.

For all experiments, each LLM is instructed to:
(i) Provide examples, in a selected domain, of causal variables and their corresponding values, interaction entities as noun phrases.
(ii) Ensure the independence of causal variables.
(iii) Avoid vague values such as "low," "high," or "moderate" and instead generate specific values that clarify the actual variable they describe.
(iv) Evaluate positive and negative examples.

\subsubsection*{Experiment on Task~1: Distilling Causal Domain Knowledge into Causal Variables} 
Selected LLMs are instructed to provide examples of causal variables and their corresponding values.
The dataset is sampled for positive and negative examples. 
Positive examples include two entities under the same causal variable, while negative examples include entities from different ones.
The same LLMs receive these examples and determine if the entities represent different values of the same causal variable. 
The LLMs are instructed to generate the causal variable name in cases where the entities represent different values of the same causal.

\subsubsection*{Experiment on Task~2: Detecting Interaction Entities} 
Selected LLMs are instructed to provide examples of interaction entities and their corresponding causal variables.
The dataset is sampled for positive and negative examples.
Positive examples include interaction entities and their corresponding causal variables, while negative examples include interaction entities with different causal variables.
The same  LLMs receive these examples and determine if the interaction entities represent values of the provided causal variables simultaneously.
If they do, the LLMs generate the causal variable values. 

\subsubsection*{Evaluation Measure}
Both tasks can be formulated as binary classification. Hence, as score functions, we use the 
precision ($P$), the recall ($R$), 
the harmonic mean of the precision and recall $F_1$ score and Cohen's $kappa$ coefficient.

\section{Results}

In the context of our experiment, we have chosen the following domains:
(i) Computer Science (Com.Sci.), (ii) Finance, (iii) Health, (iv) Physics, 
(v) Psychology, (vi) Semiconductor Manufacturing (Semi.Man.),  Sociology,  and  (vii) Urban Studies (Urban S.). 
Also, we have selected the following LLMs to be tested: 1)  GPT-4-turbo~\citep{achiam2023GPT}, 2) GPT-3.5-turbo, 3) Llama3-70b~\citep{touvron2023llama}, 4) Llama3-8b, 5) Mixtral-8×22b, 6) Mixtral-8×7b~\citep{jiang2024mixtral}, 7) Mistral-7b~\citep{jiang2023mistral}.
GPT-4-turbo and GPT-3.5-turbo LLMs experiments were conducted using the official OpenAI API\footnote{https://api.openai.com/v1.}. 
For the remaining LLMs, we utilized a third-party API called LLAMA API\footnote{https://api.llama-api.com}.

In the context of the experiment on Task~1, each LLM was instructed to generate ten examples of causal variables along with their realized values for each domain.
These examples were sampled to generate 560 positive examples, where two entities belong to the same causal variable, and 560 negative examples, where two entities belong to different causal variables. Each LLM was then tasked with predicting all the positive and negative examples. 
The results, reported in Table~\ref{tab:Task1 LLMs perfromence}, indicate that all LLMs tend to have higher precision than recall. The best-performing LLMs achieve a Cohen's $kappa$ coefficient of 35\%-40\%, which can be interpreted as fair agreement, highlighting the task difficulty. 
LLMs that use sparse expert models, such as Mixtral-8×22b, tend to perform poorly on this task. LLMs with more parameters, such as GPT-4-turbo, Llama3-70b, and Mixtral-8×22b, outperform their smaller counterparts. 
There is a significant difference among smaller LLMs, with Llama3-8b performing almost ten times better than other LLMs in the same category.

\begin{table}
  \centering
  \footnotesize
\caption{\textbf{LLMs Performance on Task~1: Distilling Causal Domain Knowledge into Causal Variables.}
  LLMs with more parameters perform better than their versions with  fewer parameters.
  Llama3-8b performs almost ten times better than LLMs in the same category.}
  \begin{tabular}{ l r r r r }
    \toprule
    \textbf{LLM} & \textbf{$F_1$ } & \textbf{$P$} & \textbf{$R$} & \textbf{ $Kappa$} \\ 
    \midrule
    GPT-4-turbo & 58\% & \textbf{97\%}  & 41\%  & \textbf{40\%} \\ 
    GPT-3.5-turbo & 6\%  & 76\% & 3\%  & 2\%  \\ 
    Llama3-70b & \textbf{59\%}  & 91\%  & \textbf{43\%}  & 39\%  \\ 
    Llama3-8b & 56\%  & 86\%  & 42\%  & \textit{35\%} \\ 
    Mixtral-8×22b & 32\%  & 96\%  & 19\%  & 18\%  \\ 
    Mixtral-8×7b & 6\%  & 94\%  & 3\%  & 3\%  \\ 
    Mistral-7b & 8\%  & 96\%  & 4\%  & 4\%  \\ 
    \bottomrule
  \end{tabular}

  \label{tab:Task1 LLMs perfromence}
\end{table}

In the context of the experiment on Task~2, each LLM was instructed to generate ten examples of interaction entities along with their corresponding causal variables and values for each domain. 
These examples were sampled to generate 559 positive examples, in which an interaction entity is sampled along with the different causal variables it represents, and 560 negative examples, in which an interaction entity is sampled along with the different causal variables selected randomly. 
Each LLM was then tasked with predicting all the positive and negative examples. 
The results, summarized in Table~\ref{tab:Task2 LLMs  perfromence}, show that all LLMs tend to have higher precision than recall. 
The best performing LLMs achieve a Cohen's $kappa$ coefficient of 47\%-60\%, which can be interpreted as moderate agreement, signifying that Task~2 is simpler than Task~1. 
LLMs that use sparse expert models, such as Mixtral based LLMs, tend to perform better on this task. 
LLMs with more parameters, such as GPT-4-turbo, Llama3-70b, and Mixtral-8×22b, outperform their smaller counterparts. 
There is a significant difference among smaller LLMs, with Llama3-8b performing almost twice as well as other LLMs in the same category.

\begin{table}
  \centering
  \footnotesize
    \caption{\textbf{LLMs performance on Task~2: Detecting Interaction Entities.} 
  LLMs that use sparse expert models, such as Mixtral-8×22b, tend to perform better on this task.}
  \begin{tabular}{ l c c c c }
    \toprule
    \textbf{LLM} & \textbf{$F_1$ } & \textbf{$P$} & \textbf{$R$} & \textbf{ $Kappa$} \\ 
    \midrule
    GPT-4-turbo & 51\%  & \textbf{95\% } & 35\%  & 33\%  \\ 
    GPT-3.5-turbo & 40\%  & 94\%  & 26\%  & 24\%  \\
    Llama3-70b & 67\%  & 90\%  & 53\%  & 47\%  \\ 
    Llama3-8b & 66\%  & 81\%  & 55\%  & \textit{42\% } \\
    Mixtral-8×22b & \textbf{77\%}  & 92\%  & \textbf{65\% } & \textbf{60\% } \\
    Mixtral-8×7b  & 21\%  & 93\%  & 12\%  & 11\%  \\
    Mistral-7b & 8\%  & 89\%  & 4\%  & 4\%  \\
    \bottomrule
  \end{tabular}

  \label{tab:Task2 LLMs perfromence}
\end{table}

\subsection*{Investigating domain dependency}

In this study, the dependencies between the selected domain and the performance of the LLMs are investigated to understand how domain-specific factors influence model effectiveness.
The dependencies between the selected domain and the performance of the LLMs are observed by stratifying the LLMs performance based on the selected domain. 
The results for Task~1, as shown in Table~\ref{tab:Task1_model_performance_on_domain}, indicate that all LLMs, except Llama3-8b, perform significantly better on data generated related to the Health domain for Task~1 compared to other domains.

\begin{table}
  \centering
  \footnotesize
    \caption{\textbf{Comparison of LLMs performance on Task~1 across different domains.}
  The agreement for the data generated by LLMs about the health domain is significantly higher than for other domains.}
  \begin{tabular}{l c c c c}
    \toprule
    \multirow{2}{*}{Domain} &\multicolumn{4}{c}{$Kappa$} \\
    & GPT-4 turbo & Llama3-70b & Mixtral-8×22b & Llama3-8b \\
    \midrule
    Com.Sci. & 27\%  & 14\% & 9\% & 23\%  \\
    Finance & 44\%  & 46\%  & 21\%  & 31\%  \\
    Health & \textbf{67\% } & \textbf{59\% } & \textbf{37\% } & \textit{46\% } \\ 
    Physics & 17\%  & 11\%  & 9\%  & 39\%  \\ 
    Psychology & 26\%  & 43\%  & 16\%  & 31\%  \\
    Semi.Man.& 46\%  & 54\%  & 21\%  & \textbf{50\% }\\
    Sociology & 53\%  & 41\%  & 14\%  & 36\%  \\ 
    Urban S. & 41\% & 43\%  & 20\%  & 26\%  \\ 
    \bottomrule
  \end{tabular}

  \label{tab:Task1_model_performance_on_domain}
\end{table}

Similarly, the results for Task~2, as shown in Table~\ref{tab:Task2_model_performance_on_domain}, indicate that all LLMs, except Mixtral-8×22b, perform better on data generated about the Health domain for Task~2 compared to other domains. 
Surprisingly, Mixtral-8×22b performs better across all the selected domains than the other tested LLMs.

\begin{table}
  \centering
  \footnotesize
    \caption{\textbf{Comparison of LLMs performance on Task~2 across different domains.} Mixtral-8×22b achieves higher performance across all the selected domains compared to the other tested LLMs}
  \begin{tabular}{l c c c c}
    \toprule
    \multirow{2}{*}{Domain} &\multicolumn{4}{c}{$Kappa$} \\
    & GPT-4 turbo & Llama3-70b & Mixtral-8×22b & Llama3-8b \\ 
    \midrule
    Com.Sci. & 28\%  & 46\%  & 54\%  & 38\%  \\ 
    Finance & 23\%  & 39\%  & 41\%  & 27\%  \\
    Health & \textbf{53\% } & \textbf{56\%  }& 69\%  & \textbf{59\% } \\ 
    Physics & 49\%  & 55\%  & 71\%  & 56\%  \\
    Psychology & 34\%  & 47\%  & 54\%  & 34\%  \\ 
    Semi.Man.  & 26\%  & 36\%  & 41\%  & 34\%  \\ 
    Sociology & 37\%  & 54\%  & 76\%  & 55\%  \\ 
    Urban S.  & 17\%  & 46\%  & \textbf{74\% } & 36\%  \\ 
    \bottomrule
  \end{tabular}

  \label{tab:Task2_model_performance_on_domain}
\end{table}

\subsection*{Investigating the dependency between the generated data and LLMs performance}

The dependencies between the generated data and the LLMs performance are explored by utilizing a text embedding model\footnote{Text-embedding-3-large model provided by OpenAI}.
The text embedding model is used to extract the embedding vectors of the inputs used to prompt LLMs. 
Specifically, the cosine similarity is calculated between the embedding vectors of the two entities for Task~1 and between the interaction entity embedding vector and the embedding vector of the provided causal variables for Task~2.
The calculated cosine similarity is then used for a classification method based on a similarity threshold. 
Higher agreement with the classification results at a lower cosine similarity threshold indicates that the generated or the predicted data is more dissimilar.
Conversely, higher agreement with the classification results at a higher cosine similarity threshold indicates that the generated or the predicted data is more similar.

\begin{figure}
\centering
  \includegraphics[width=0.5\columnwidth]{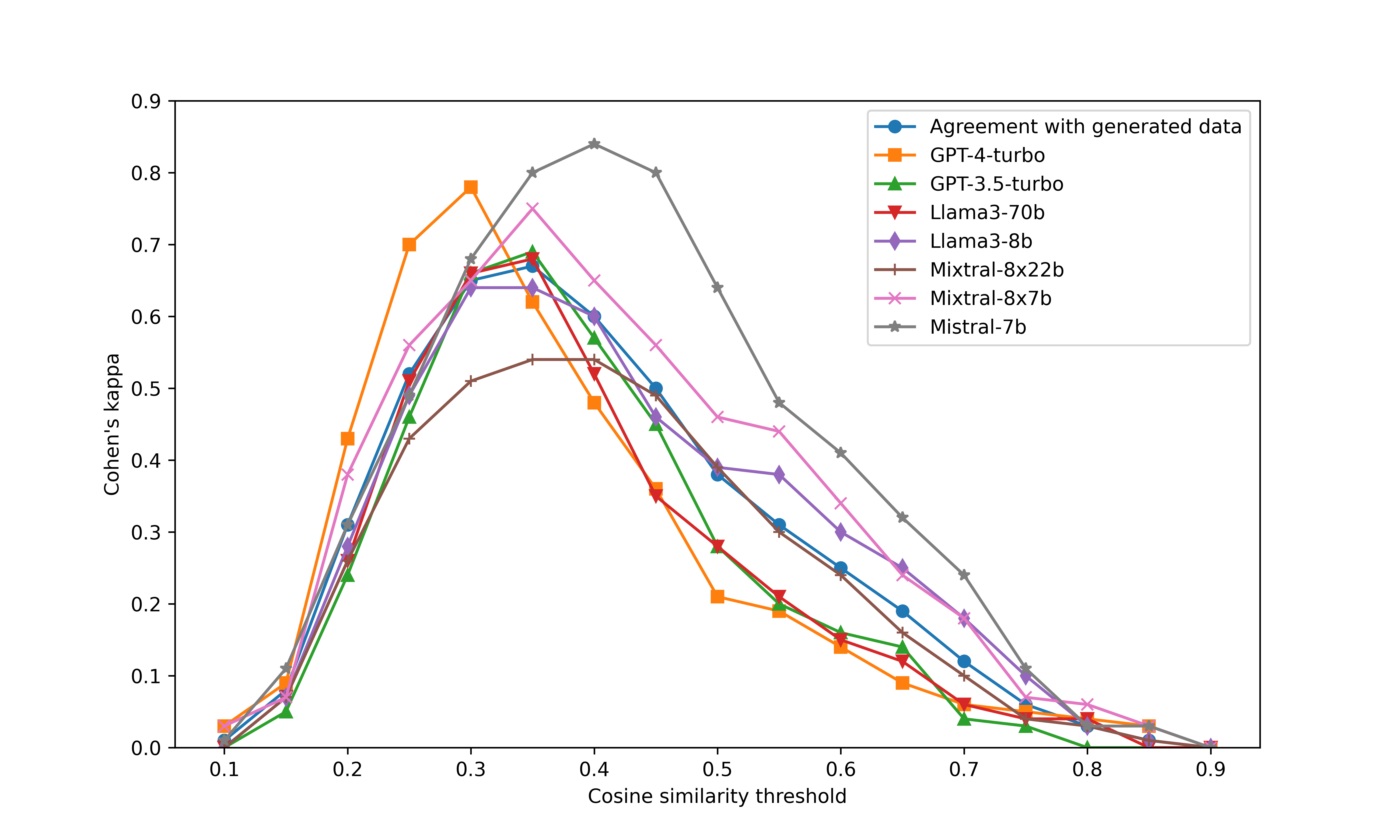}
  \caption{\textbf{The agreement between the generated data for Task~1 and the classification results based on the cosine similarity threshold.} Agreement at a lower cosine similarity threshold indicates that, in the generated data, two entities represent different values of the same causal variable and are not necessarily semantically similar.
  The data generated by Mixtral-8×22b tends to have a higher cosine similarity to other LLMs.}
  \label{fig:cos sim Task1 left}
\end{figure}

For the case of Task~1, as shown in Figure~\ref{fig:cos sim Task1 left}, agreement between generated data and the classification results at a lower cosine similarity threshold suggests that the generated two entities represent different values of the same causal variable and are not necessarily semantically similar.
Also, it can be seen that the data generated by Mixtral-8×22b tends to show higher agreement at higher similarity thresholds compared to other LLMs.

\begin{figure}
\centering
  \includegraphics[width=0.5\columnwidth]{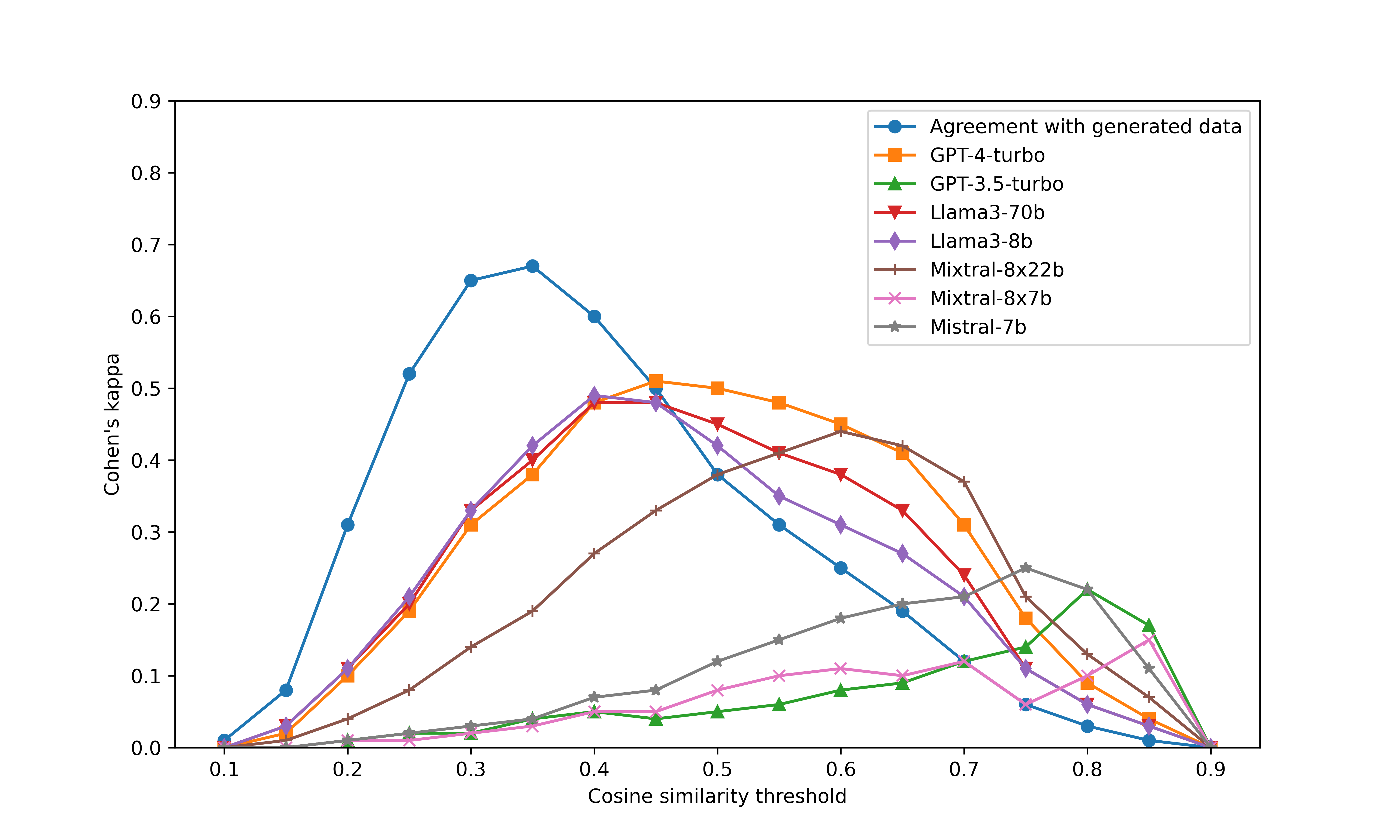}
  \caption{\textbf{The agreement between LLMs predictions and the classification is based on the cosine similarity threshold.}
  GPT-3.5-turbo, Mixtral-8×7b and Mistral-7b reach their peak agreement at a higher threshold, less aligned with the generated data.
  GPT-4-turbo, Llama3-70b and Llama3-8b reach their peak agreement at lower similarity thresholds, which aligns more with the generated data.}
  \label{fig:cos sim Task1 right}
\end{figure}

Moreover, as shown in Figure~\ref{fig:cos sim Task1 right}, higher agreement between the classification results and the LLMs predictions on Task~1 at higher similarity thresholds indicates that LLMs predict two entities as different values of the same causal variable if they are semantically similar. 
GPT-3.5-turbo, Mixtral-8×7b and Mistral-7b reach their highest agreement at a higher threshold, which aligns less closely with the generated data.
In contrast, GPT-4-turbo, Llama3-70b and Llama3-8b achieve their highest agreement at lower similarity thresholds, which aligns more closely with the generated data.

\begin{figure}
\centering
  \includegraphics[width=0.5\columnwidth]{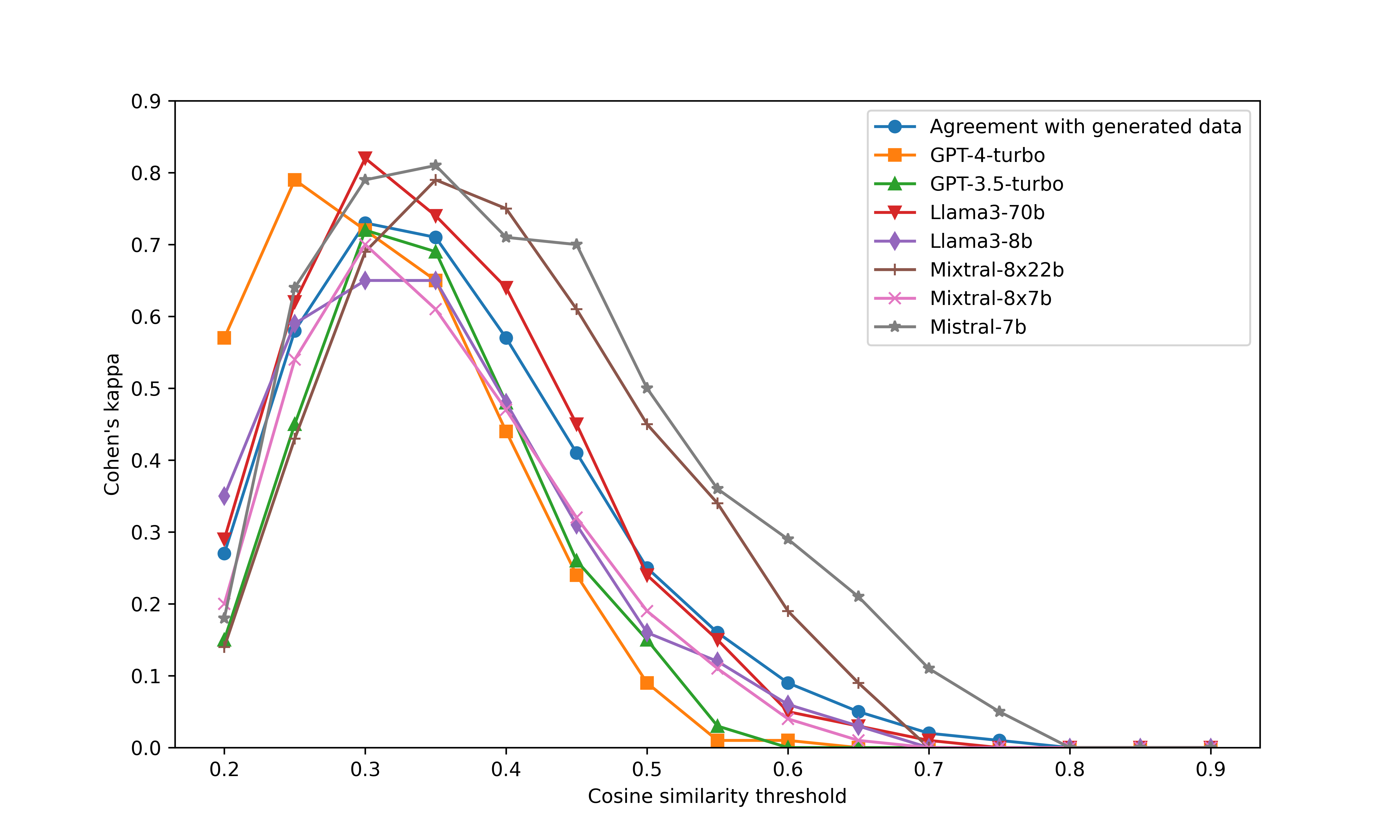}
  \caption{\textbf{The agreement between the generated data for Task~2 and the classification results based on the cosine similarity threshold.} Agreement at a lower cosine similarity threshold indicates that a text represents values of different causal variables and is not necessarily semantically similar to these variables.
  The data generated by Mixtral-8×22b tends to exhibit higher agreement at higher similarity thresholds compared to other LLMs.}
  \label{fig:cos sim Task2 left}
\end{figure}

For the case of Task~2, the results, as shown in Figure~\ref{fig:cos sim Task2 left}, show high agreement between the generated data and the classification results at a lower cosine similarity threshold.
Hence, this indicates that a text entity representing values of different causal variables is not necessarily semantically similar to the text entities representing these different causal variables. 
Also, the data generated by Mixtral-8×22b tends to exhibit higher agreement at higher similarity thresholds compared to other LLMs.

\begin{figure}
\centering
  \includegraphics[width=0.5\columnwidth]{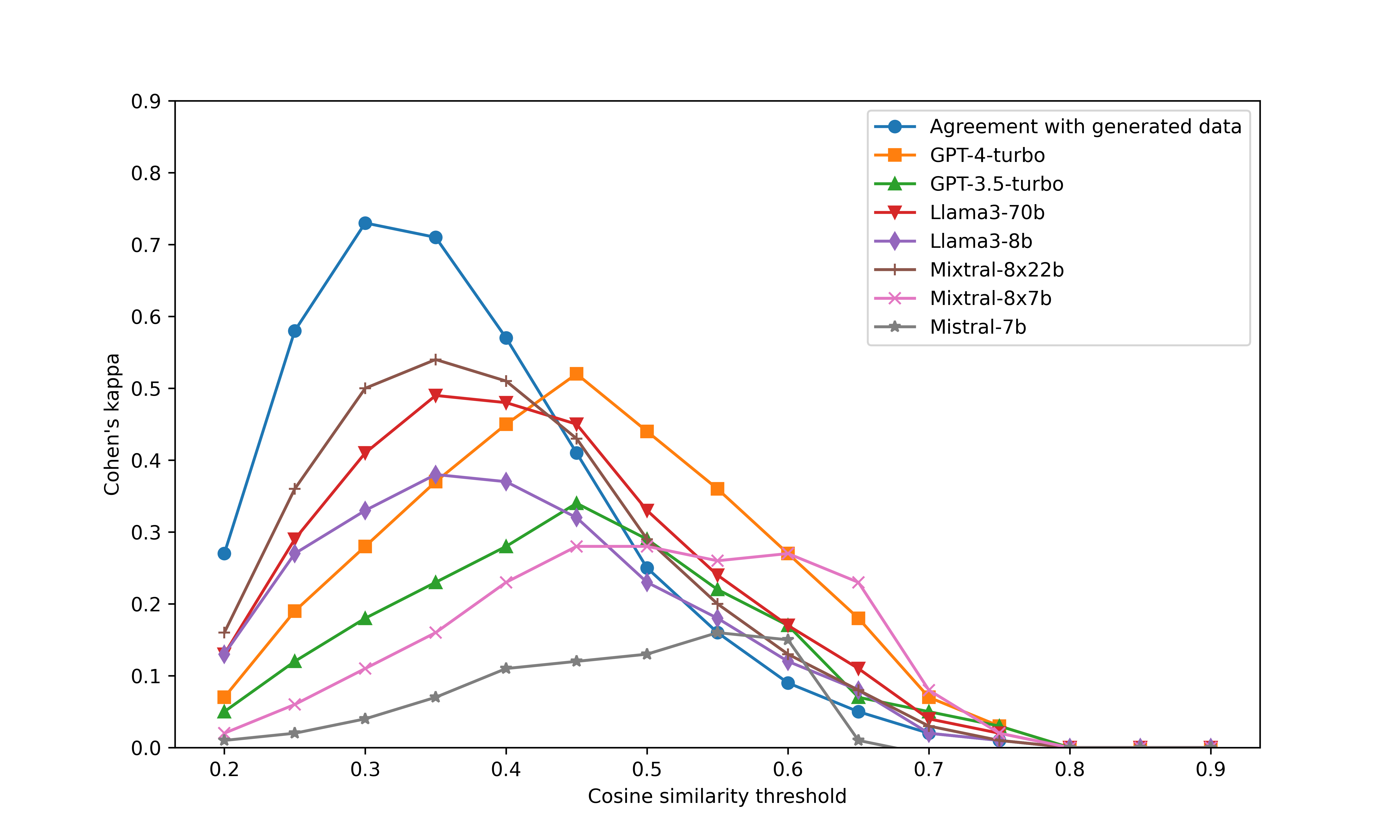}
  \caption{\textbf{ The agreement between the LLMs predictions on Task~2 and the classification results based on the cosine similarity threshold.} Higher agreement with the model predicting at higher similarity thresholds indicates that LLMs, to a certain extent, predict a text as values of the different causal variables if they are semantically similar.
Llama3-70b and Mixtral-8×22b reach their peak agreement at lower similarity thresholds, which aligns more with the generated data.
GPT-3.5-turbo and GPT-4-turbo reach their maximum agreement at a higher threshold, less in line with the generated data.}
  \label{fig:cos sim Task2 right}
\end{figure}

Moreover, higher agreement between the the LLMs predicting on Task~2 and the classification results at higher similarity thresholds, as shown in Figure~\ref{fig:cos sim Task2 right}, indicates that LLMs tend to predict a text as values of different causal variables if they are semantically similar. 
GPT-3.5-turbo and GPT-4-turbo achieve their maximum agreement at a higher threshold, which aligns less closely with the generated data.
In contrast, Llama3-70b and Mixtral-8×22b reach their peak agreement at lower similarity thresholds, which aligns more closely with the generated data. 

\section{Discussion}

The results of  this study highlight several important findings related to understanding LLMs causal modeling abilities.
Firstly, different LLMs perform differently on each task.
This indicates the effectiveness of the proposed two tasks for comparing LLMs causal modeling abilities.
Specifically, GPT-4-turbo, Llama3-70b and Llama3-8b achieve higher agreement in Task~1: distilling causal domain knowledge into causal variables, while Mixtral-8×22b achieves higher agreement on Task~2: detecting interaction entities.
Secondly, all LLMs tend to achieve significantly higher precision than recall, which may be attributed to the Temperature parameter being set to 0. This pushes the LLMs production towards more deterministic output and avoids prediction with lower confidence.
Thirdly, all LLMs generate examples for Task~1 and Task~2 which have high agreement with the classification results based on the cosine similarity calculated on the embedding vector of the text entities.
This highlights that text embedding models are not useful for addressing Task~1 and Task~2 using classification methods based on a cosine similarity threshold out of the box. 
Fourthly, all LLMs, with the exception of GPT-4-turbo, exhibit better performance on Task~2 than on Task~1. This can be attributed to the difficulty of Task~1 where the causal variable is not given and needs to be inferred from the provided text entities.
As such, Task~1 is more knowledge intensive than Task~2 where the causal variables are given alongside the text entity.
Our study supports this interpretation by  revealing  that all LLMs have a tendency to predict two text entities belonging to the same causal variable when the entities have higher cosine similarity between their embedding vectors, as shown in Figure~\ref{fig:cos sim Task1 right}. 
This trend is less dominant for Task~2 as shown in Figure~\ref{fig:cos sim Task2 right}.
Finally, the study unveils a significant dependency between the domain from which the data is generated and the LLMs prediction performance. 
For instance, in Table~\ref{tab:Task1_model_performance_on_domain},  instances related to health are predicted more accurately for all LLMs except Llama3-8b, indicating a strong connection between the domain and the LLMs performance on Task~1. 
The superior performance of LLMs in the Health domain can be attributed to their training data being more closely related to that domain, as well as the strict standards for causal language and variable definition within the Health domain.
However, comparing the results of Llama3-70b and GPT-4-turbo with those of Llama3-8b  in Table\ref{tab:Task1_model_performance_on_domain} reveals that LLMs tend to overfit those domains and lose the ability to extend it to other domains such as Physics.
This dependency is less apparent for Task~2, as the performance of the LLMs is comparable across several domains.

Our study builds on existing research from the field of causality to model causal variables \citep{suzuki2020causal,Kuorikoski}, and interaction entities~\citep{mcdonnell2018transitivity,VanderWeele_2009} by formulating the modeling challenges for domain knowledge articulated as relations between text entities in Section~\ref{Problem formulation}.
The study complements recent research on the test of LLM causal ability~\citep{jin2023can,zevcevic2023causal}.
\citealp{jin2023can} tested the abilities of LLMs in identifying  different types of relation between causal variables given a text description of a causal model.
Our study, in fact, addresses more fundamental causal ability on how to transfer domain knowledge described in text entities to causal variables, which is intuitively more natural to be addressed using LLMs.
The findings of this study are also in line with \citealp{zevcevic2023causal} as we argue that for knowledge intensive tasks, such as Task~1, LLMs performance highly depends on the domain in which the data are generated, as the LLMs tend to only recite what it was trained on.

The presented study demonstrates that while LLMs are not flawless and currently undergo huge improvements, they can already  be beneficial in causal modeling. The implications of this study are multifaceted, ranging from aiding domain experts in enhancing the modeling of causal domain knowledge as documented in cognitive maps, such as those used in urban blight analysis~\citep{pinto2023analyzing}, to utilizing LLMs in post processing the results of causal information extraction methods, as exemplified by~\cite{gopalakrishnan2023text}. 
Specifically, the study findings underscore the need for thorough analysis when selecting the appropriate LLM to achieve optimal results.
Consequently, this study lays the groundwork for a more effective transfer of knowledge from the causal domain into representations that align with recommendations from the field of causal data science.
This, in turn, could lead to significant advancements in natural language processing technology, potentially resulting in more robust and reliable data analysis with reduced bias.

\textbf{Limitations:}
This study significantly narrow down the search space for the appropriate LLM for causal modeling.
However, these exact numbers have been derived based on generated data. 
Therefore, for a refined analysis, the research findings need to be compared with the expertise of domain-specific experts.
In addition, it is important to note that our study employs the same prompt in the same language (i.e. English)  for all LLMs under investigation.
This opens up opportunities for further research into the relationship between the prompt, the used language and the causal modeling capabilities of LLMs.

\section{Conclusion}

In this work, we introduced two novel tasks to transform causal domain knowledge into a representation that aligns more closely with guidelines from causal data sciences.
Task~1 is focused on identifying whether two text entities represent different values of the same causal variable. At the same time, Task~2 is concerned with determining if a text entity represents values of different causal variables. 
We evaluated an extensive list of LLMs on these new tasks and showed that off-the-shelf LLMs perform fairly well. 
Our experiments show significant improvements in the abilities of LLMs in domain knowledge causal modeling compared to previous generations, such as GPT-3.5-turbo.
Our findings indicate that Task~2 is simpler since it assumes that the causal variables are provided. 
Additionally, for Task~2, sparse expert models such as Mixtral-8×22b  demonstrate better performance.
Regarding Task~1, our findings suggest that larger LLMs such as Llama3-70b and GPT-4-turbo exhibit better performance as they can encapsulate more knowledge from the data on which they were initially trained.
However, they are prone to overfitting to specific domains, reducing their ability to generalize to other domains.

\bibliographystyle{unsrtnat}
\bibliography{references} 
\end{document}